\documentclass[runningheads]{llncs}
\usepackage[T1]{fontenc}
\usepackage[hyphens]{url}  
\usepackage{graphicx} 
\usepackage{caption} 

\usepackage{soul}
\usepackage{url}
\usepackage[utf8]{inputenc}
\usepackage{amsmath}
\usepackage{booktabs}
\usepackage{algorithm}
\usepackage{algorithmic}
\usepackage{multirow}
\usepackage{amssymb}
\usepackage{subcaption}
\usepackage{enumitem}
\usepackage{bbm}
\usepackage{hhline}
\usepackage{subcaption}

\newcommand{\CF}{\textrm{cf}}

\usepackage{color}

\title{Achieving Counterfactual Fairness for Anomaly Detection}

\author{Xiao Han\inst{1} \and
Lu Zhang \inst{2} \and
Yongkai Wu \inst{3} \and
Shuhan Yuan \inst{1} 
}

\authorrunning{X. Han et al.}

\institute{
Utah State University, Logan, UT 84322, USA \\ \email{\{xiao.han,shuhan.yuan\}@usu.edu}\and
University of Arkansas, Fayetteville, AR 72701, USA \\ \email{lz006@uark.edu} \and
Clemson University, Clemson, SC 29634, USA \\ \email{yongkaw@clemson.edu }
}

\begin{document}

\maketitle

\begin{abstract}
    Ensuring fairness in anomaly detection models has received much attention recently as many anomaly detection applications involve human beings. 
    However, existing fair anomaly detection approaches mainly focus on association-based fairness notions. In this work, we target counterfactual fairness, which is a prevalent causation-based fairness notion. The goal of counterfactually fair anomaly detection is to ensure that the detection outcome of an individual in the factual world is the same as that in the counterfactual world where the individual had belonged to a different group. To this end, we propose a counterfactually fair anomaly detection (CFAD) framework which consists of two phases, counterfactual data generation and fair anomaly detection.
    Experimental results on a synthetic dataset and two real datasets show that CFAD can effectively detect anomalies as well as ensure counterfactual fairness.
    \keywords{Anomaly Detection  \and Counterfactual Fairness.}
\end{abstract}

\section{Introduction}

Anomaly detection, which aims to detect samples that are deviated from the normal ones, has a wide spectrum of applications, such as transaction fraud detection and cyber-intrusion detection. In recent years, deep anomaly detection models, powered by complex deep neural nets, have made promising progress in effectively detecting anomalies.
Besides effectiveness, researchers recently notice the importance of taking the societal impact of anomaly detection into consideration as many anomaly detection tasks involve human individuals. Fairness as one fundamental component to build trustworthy AI has received much attention. 
Recent studies have shown that anomaly detection models can incur discrimination against certain groups. For example, a deep anomaly detection model could overly flag black males as anomalies \cite{zhangFairDeepAnomaly2021a}. In the scenarios of credit risk analysis, anomaly detection models predict more females as anomalies \cite{songDeepClusteringBased2021}. 

Several fair anomaly detection models have been proposed, which ensure no discrimination against a particular group based on the sensitive feature \cite{pFairOutlierDetection2020,zhangFairDeepAnomaly2021a,songDeepClusteringBased2021,shekharFairODFairnessawareOutlier2021,almanza2022k}.
However, these approaches mainly focus on achieving association-based fairness notions like demographic parity. Recent studies have demonstrated the importance of treating fairness as causation-based notions that concern the causal effect of the sensitive feature on the model outcomes \cite{kilbertus2018avoiding,nabi2018fair,van2021decaf}. Counterfactual fairness is one important causation-based fairness notion \cite{kusnerCounterfactualFairness2018}. It considers that a model is fair if for a particular individual the model outcome in the factual world is the same as that in the counterfactual world where the individual had belonged to a different group. To the best of our knowledge, no studies have been conducted to ensure counterfactual fairness in anomaly detection.

In this work, we focus on counterfactual fairness for anomaly detection with the goal to ensure that the detection outcomes remain consistent in both the factual and counterfactual worlds. 
Achieving counterfactual fairness for anomaly detection is challenging. First, we can only observe the factual data. The counterfactual data are unobservable and cannot be obtained by simply changing the sensitive feature of the factual data. This is because the data generation is governed by an underlying causal mechanism where any intervention on one feature will subsequently affect the values of other features. 
Second, in anomaly detection, we can only observe factual normal data. How to build a detection model which ensures the detection results be unchanged for individuals across the factual and counterfactual worlds while also preserving high anomaly detection performance imposes additional challenges.

To tackle the above challenges, we propose a Counterfactually Fair Anomaly Detection (CFAD) framework. We do not require the knowledge of the causal graph and structural equations but only assume that the data generation follows a generalized linear SCM.
We use an autoencoder as the base anomaly detection model where the anomaly score of a sample is derived based on the reconstruction error of the autoencoder. 
Then, we propose a two-phase approach.
In the first phase, motivated by \cite{ngGraphAutoencoderApproach2019} which leverages the graph autoencoder for causal structure learning from observed data, we develop an approach to generate counterfactual data based on a graph autoencoder.
In the second phase, we apply adversarial training \cite{edwardsCensoringRepresentationsAdversary2016,madrasLearningAdversariallyFair2018} on a vanilla autoencoder to achieve counterfactual fairness for anomaly detection. The idea is to ensure that the hidden representations of factual and counterfactual data derived from the encoder cannot be distinguished by a discriminator. As a result, the reconstruction error, i.e., anomaly score, will not differ much between the factual and counterfactual data, leading to similar detection results for both factual and counterfactual data.

\section{Preliminary}
{\bf \noindent Structural Causal Model (SCM).}
Our work adopts Pearl's Structural Causal Model (SCM) \cite{pearlCausality2009} as the prime methodology for defining and measuring counterfactual fairness.
Throughout this paper, we use the upper/lower case alphabet to represent variables/values.

\begin{definition}
	An SCM is a triple $\mathcal{M}= \{U,V,F\}$ where 
	\begin{itemize}[leftmargin=*]
	    \item[ 1)] $U$ is a set of exogenous variables that are determined by factors outside the model. A joint probability distribution $P(u)$ is defined over the variables in $U$.
	    \item[ 2)] $V$ is a set of endogenous variables that are determined by variables in $U\cup V$.
	    \item[ 3)] $F$ is a set of deterministic functions $\{f_1,\ldots,f_n\}$; for each $X_i \in V $, a corresponding function $f_{i}$ is a mapping from $U \cup (V\setminus \{X_i\})$ to $X_i$, i.e., $X_{i}=f_{i}(X_{pa(i)},U_{i})$, where $X_{pa(i)}\subseteq V\backslash \{X_{i}\}$ called the parents of $X_{i}$, and $U_{i}\subseteq U$.
	\end{itemize}
\end{definition}

An SCM is often illustrated by a causal graph $\mathcal{G}$ where each observed variable is represented by a node, and the causal relationships are represented by directed edges $\rightarrow$. In this graphical representation, the definition of parents is consistent with that in the SCM. 

Inferring causal effects in the SCM is facilitated by the do-operator which simulates the physical interventions that force some variable $X\in V$ to take a certain value $x$. 
For an SCM $\mathcal{M}$, intervention $\textrm{do}(X=x)$ is equivalent to replacing original function in $F$ with $X = x$. 
After the replacement, the distributions of all variables that are the descendants of $X$ may be changed. We call the SCM after the intervention the submodel, denoted by $\mathcal{M}[x]$. For any
variable $Y\in V$ which is affected by the intervention, its interventional
variant in submodel $M[x]$ is denoted by $Y[x]$. 

{\bf \noindent Counterfactuals.} 
Counterfactuals are about answering questions such as for two variables $X,Y\in V$, whether $Y$ would be $y$ had $X$ been $x$ in unit (or situation) $U = u$.
Such question involves two worlds, the factual world represented by $\mathcal{M}$ and the counterfactual world represented by $\mathcal{M}[x]$, and hence cannot be answered directly by the do-operator.  
When the complete knowledge of the SCM is known, the counterfactual quantity can be computed by the three-step process:
\begin{itemize}[leftmargin=*]
\item[ 1)] Abduction: Update $P(u)$ by evidence $e$ to obtain $P(u|e)$.
\item[ 2)] Action: Modify $\mathcal{M}$ by performing intervention $\textrm{do}(x)$ to obtain the
submodel $\mathcal{M}[x]$.
\item[ 3)] Prediction: Use modified submodel $\mathcal{M}[x]$ with updated probability $P(u|e)$ to
compute the probability of $Y=y$.
\end{itemize}

\section{Counterfactually Fair Anomaly Detection}

\subsection{Counterfactual Fairness}
We start by defining counterfactual fairness in the context of anomaly detection. Following the typical anomaly detection setting, we assume a training set $\mathcal{D}=\{d^{(n)}\}_{n=1}^N$ which consists of $N$ normal samples/individuals and a test set that consists of both normal samples and anomalies. Each sample is given by $d^{(n)}=\{s^{(n)},x^{(n)}\}$ where $S$ denotes a binary sensitive variable and $X=\{X_i ~|~ i=1:m\}$ denotes all other variables (i.e., profile attributes). We then use $Y$ to denote the anomaly label. For representation, we use $S=\{s^{+},s^{-}\}$ to denote advantage and disadvantage groups respectively, and use $Y=\{0,1\}$ to denote normal samples and anomalies respectively.
Our goal is to learn a detection model for computing an anomaly score $g(x^{(n)})$ based on the profile attributes for each individual $n$ which can be used to judge whether it is a normal sample or an anomaly.

To define counterfactual fairness, similar to \cite{kusnerCounterfactualFairness2018}, for each individual $d^{(n)}$ we consider its instance in the counterfactual world $\mathcal{M}_{s}$ by flipping the value of its sensitive variable to the opposite $s$ (i.e., $s^{+}$ becomes $s^{-}$ and vice versa), denoted by $d^{(n)}_{\CF}=\{s,x^{(n)}_{\CF}\}$ where $x^{(n)}_{\CF}$ represents the profile attributes in the counterfactual world. Note that $x^{(n)}_{\CF}$ may not be the same as $x^{(n)}$ due to the causal relation between $S$ and $X$ in the underlying data generation mechanism. Then, counterfactual fairness is defined as:
\begin{definition}
    An anomaly detection model is counterfactually fair if for each individual $n$ we have $g(x^{(n)})=g(x^{(n)}_{\CF})$.
\end{definition}

\subsection{Overview of Counterfactually Fair Anomaly Detection (CFAD)}

\begin{figure}[t]
\centering
\includegraphics[width=.95\textwidth]{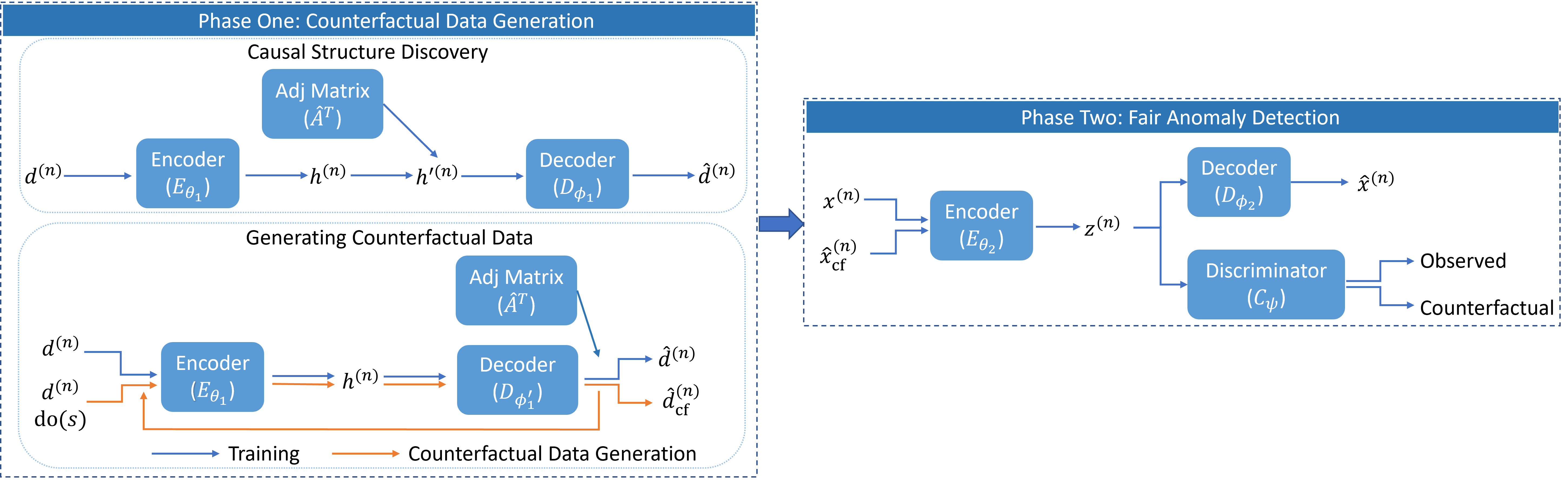}
\caption{Framework of CFAD}
\label{fig:cfad}
\end{figure}

The goal of CFAD is to train an anomaly detection model on $\mathcal{D}$ that can: (1) effectively detect anomalies, and (2) ensure counterfactual fairness. To achieve this goal, CFAD consists of two phases, counterfactual data generation and fair anomaly detection. Counterfactual data generation is to generate a counterfactual dataset $\mathcal{D}_{\CF}=\{d^{(n)}_{\CF}\}_{n=1}^N$ of $\mathcal{D}$ in which each counterfactual sample is generated by the submodel which flips the value of the sensitive variable to its counterpart. To this end, we assume a generalized linear SCM and develop a novel graph autoencoder for the data generation. In the second phase, we make use of a standard autoencoder for anomaly detection where the anomaly score is derived based on the reconstruction error. To achieve fairness, we develop an adversarial training framework to train the autoencoder by taking the factual and counterfactual data as inputs. The idea is to make the hidden representations of the autoencoder not encode the information of the sensitive variable so that intervening the sensitive variable would not change the detection outcome. Figure \ref{fig:cfad} shows the framework of CFAD.

\subsection{Phase One: Counterfactual Data Generation}
We assume that the data generation follows a generalized linear SCM, which is a common assumption in gradient-based causal discovery \cite{vowels2021d}. 
To ease representation, we also assume that $S$ has no parents in the SCM. Our method can easily extend to cases where $S$ has parents by keeping the values of $S$'s parents unchanged in the counterfactual world since the intervention on $S$ has no influence on its parents. Thus, W.L.O.G. the structural equation of each variable $X_i$ in $X$ can be written as follows.
\begin{equation}\label{eq:glm}
    X_i = A_{1,i}\cdot f(S) + \sum_{X_j \in X_{pa(i)}\backslash\{S\}} A_{j,i} \cdot f(X_j) + U_i,
\end{equation}
where $f(\cdot)$ can be any linear/nonlinear function and $A_{j,i}$ is an element in the adjacency matrix $A\in \mathbb{R}^{(m+1)\times (m+1)}$ which indicates the weights of the generalized linear SCM. Each sample $d^{(n)}=\{s^{(n)}, \{x^{(n)}_{i} ~|~ i=1:m\} \}$ satisfies Eq.~\eqref{eq:glm}. Following the Abduction-Action-Prediction process, from Eq.~\eqref{eq:glm}, we have
\begin{equation*}
    u^{(n)}_{i} = x^{(n)}_{i} -A_{1,i}\cdot f(s^{(n)}) - \sum_{X_j \in X_{pa(i)}\backslash\{S\}} A_{j,i} \cdot f(x^{(n)}_j).
\end{equation*}

Meanwhile, by performing intervention to flip $s^{(n)}$ to its counterpart $s$, the structural
equation of counterfactual variable $X_i[s]$ in the submodel $\mathcal{M}[s]$ of Eq.~\eqref{eq:glm} is given by
\begin{equation}\label{eq:glm2}
    X_i[s] = A_{1,i}\cdot f(s) + \sum_{X_j \in X_{pa(i)}\backslash\{S\}} A_{j,i} \cdot f(X_j[s]) + U_i.
\end{equation}
Note that $S$ is fixed to $s$ by the intervention and $U_i$ is not affected by the intervention. Denoting the counterfactual of $d^{(n)}$ by $d^{(n)}_{\CF}=\{s, \{x^{(n)}_{i}[s] ~|~ i=1:m\} \}$, it should satisfy Eq.~\eqref{eq:glm2}. Thus, we have
\begin{equation*}\label{eq:glm3}
    x^{(n)}_i[s] = A_{1,i}\cdot f(s) + \sum_{X_j \in X_{pa(i)}\backslash\{S\}} A_{j,i} \cdot f(x^{(n)}_j[s]) + u^{(n)}_i,
\end{equation*}
which leads to 
\begin{equation}\label{eq:glm4}
     x^{(n)}_i[s] = A_{1,i}\cdot f(s) + \!\!\!\!\!\!\! \sum_{X_j \in X_{pa(i)}\backslash\{S\}} \!\!\!\!\!\!\! A_{j,i} \cdot f(x^{(n)}_j[s]) 
     + x^{(n)}_{i} - A_{1,i}\cdot f(s^{(n)}) - \!\!\!\!\!\!\!\! \sum_{X_j \in X_{pa(i)}\backslash\{S\}} \!\!\!\!\!\!\!\! A_{j,i} \cdot f(x^{(n)}_j).
\end{equation}
Finally, we compute the value of $x^{(n)}_i[s]$ according to Eq.~\eqref{eq:glm4} following the topological order and derive $d^{(n)}_{\CF}$ from the observational data.

The challenge in the above derivation is how to estimate function $f(\cdot)$ and adjacency matrix $A$ of the SCM. Next, we develop a causal structure discovery approach based on the graph autoencoder as proposed in \cite{ngGraphAutoencoderApproach2019}.

\subsubsection{Causal Structure Discovery}
We estimate the adjacency matrix of the SCM defined in Eq.~\eqref{eq:glm} by a graph autoencoder model with parameters $\{\theta_1,\phi_1,\hat{A}\}$. Specifically, an encoder is first adopted to derive the hidden representation of a sample $d^{(n)}$, i.e., $h^{(n)} = E_{\theta_1}(d^{(n)})$, where $E_{\theta_1}(\cdot)$ is parameterized by a multilayer neural network. Then, the message passing operation is applied on the hidden representation, i.e.,  $h'^{(n)} = \hat{A}^T h^{(n)}$, where $\hat{A}$ is a parameter matrix. Finally, an decoder is used to reconstruct the original input from $h'^{(n)}$, i.e., \[\hat{d}^{(n)}=D_{\phi_1}(h'^{(n)})=D_{\phi_1}(\hat{A}^T E_{\theta_1}(d^{(n)}))\]
where $D_{\phi_1}(\cdot)$ is parameterized by a different multilayer neural network. Note that both the encoder $E_{\theta_1}(\cdot)$ and the decoder $D_{\phi_1}(\cdot)$ work in a variable-wise manner in order to preserve the order of the message passing in the SCM. To train the graph autoencoder model, the objective function is defined as:

\begin{equation*}
    \mathcal{L}_{\textrm{GAE}}(A,\theta_1,\phi_1) = \frac{1}{2N}\sum_{n=1}^N \|d^{(n)}-\hat{d}^{(n)}\|_2^2 + \lambda \|\hat{A}\|_1 
     \text{  s.t. } tr(e^{\hat{A} \odot \hat{A}})-m-1=0,
\end{equation*}
where the constraint $tr(e^{\hat{A} \odot \hat{A}})-m-1=0$ is to ensure acyclicity in the graph. After training, matrix $\hat{A}$ will be a good estimation of the adjacency matrix $A$.

One challenge in applying the graph autoencoder to our work is that, although the graph autoencoder can accurately estimate the adjacency matrix $\hat{A}$, it does not produce a good reconstruction of the input sample, which implies that it does not accurately estimate the function $f(\cdot)$ in the SCM.
In order to generate the counterfactual data, the reconstructed sample with high fidelity is critical. Hence, we improve the graph autoencoder by adding another decoder that focuses on data reconstruction, where the trained matrix $\hat{A}$ and the encoder $E_{\theta_1}(\cdot)$ are reused in this step.

In particular, we similarly feed each sample $d^{(n)}$ to trained encoder $E_{\theta_1}(\cdot)$ to obtain the corresponding hidden representation. Then, in order to be consistent with the structural equations Eq.~\eqref{eq:glm}, different from \cite{ngGraphAutoencoderApproach2019} where the message passing operation is applied in the representation space, we first use a new variable-wise decoder $D_{\phi'_1}$ to transform the hidden representation back to the original data space, and then aggregate the message from the neighbors based on matrix $\hat{A}$. As a result, the reconstruction process of each sample is given by the following equation.
\begin{equation*}
    \hat{d}^{(n)} = \hat{A}^{T} D_{\phi'_1}(E_{\theta_1}(d^{(n)})).
\end{equation*}
The objective function is to reconstruct the input with $\hat{A}$ and $\theta_{1}$ fixed:
\begin{equation*}
    \mathcal{L}_{\textrm{D}}(\phi'_1) = \frac{1}{2N}\sum_{n=1}^N \sum_{i=1}^d \|d_i^{(n)}-\hat{d}_i^{(n)}\|_2^2.
\end{equation*}
After training, we obtain the approximated mapping function $\hat{f}=D_{\phi'_1} \circ E_{\theta_1}$.

\subsubsection{Generating Counterfactual Data}
Given estimated adjacency matrix $\hat{A}$ and function $\hat{f}$, for each sample $d^{(n)}$, we generate its counterfactual $d^{(n)}_{\CF}$ following the Abduction-Action-Prediction process. We first intervene $s^{(n)}$ to its counterpart $s$ and compute $\hat{f}(s)$. Then, we sort all variables in $X$ in a topological order and compute $\hat{x}^{(n)}_i[s]$ iteratively according to Eq.~\eqref{eq:glm4} where $A$ and $f$ are replaced by their estimators $\hat{A}$ and $\hat{f}$. Finally, we obtain $\hat{D}_{\CF} = \{\hat{d}_{\CF}^{(n)}\}_{n=1}^N$, where  $\hat{d}^{(n)}_{\CF}=\{s, \{\hat{x}^{(n)}_{i}[s] ~|~ i=1:m\} \}$.

\subsection{Phase Two: Fair Anomaly Detection}
We use the autoencoder as the base model for anomaly detection, which is trained to minimize the reconstruction errors of normal samples. It is worth noting that a fully-connected autoencoder model is used here which is different from the variable-wise autoencoder used in the previous section for counterfactual data generation. Meanwhile, to achieve counterfactual fairness, we leverage the idea of adversarial training to make the hidden representations derived by the autoencoder not encode the information of the sensitive variable. To this end, we develop a pre-training and fine-tuning framework to ensure the effectiveness of anomaly detection as well as counterfactual fairness. The reason for adopting the pre-training and fine-tuning training approach instead of the end-to-end training is that some counterfactual samples in $\hat{\mathcal{D}}$ could be anomalies. If we include all samples in $\hat{\mathcal{D}}$ to train the autoencoder model, the performance of anomaly detection can be damaged. Hence, we use samples in $\mathcal{D}$ to pre-train the autoencoder model. Then, during fine-tuning, we slightly update the autoencoder so that the effectiveness of anomaly detection and the counterfactual fairness can be balanced.
Finally, we do not use the sensitive variable and only use the non-sensitive variables $X$ to train the anomaly detection model.

To be more specific, in the pre-training phase, given the training set with normal samples $\mathcal{D}$, an encoder first maps each sample $x^{(n)}$ to a hidden representation $z^{(n)}=E_{\theta_2}(x^{(n)})$, and then a decoder aims to reconstruct the original input from the hidden representation $\hat{x}^{(n)}=D_{\phi_2}(z^{(n)})$. 
The objective function is to minimize the reconstruction error of normal samples:
\begin{equation*}
    \mathcal{L}_{\textrm{AE}}(\theta_2,\phi_2) = \frac{1}{2N}\sum_{n=1}^N \|d^{(n)}- D_{\phi_2} \circ E_{\theta_2}(x^{(n)})\|_2^2.
\end{equation*}

After pre-training the autoencoder model, in order to achieve counterfactual fairness, we further incorporate the adversarial training strategy to further fine-tune the autoencoder model so that the hidden representation $z^{(n)}$ derived by the encoder is free of the information of the sensitive variable. To this end, for each sample $d^{(n)}=\{s^{(n)},x^{(n)}\}$ and its counterfactual sample $\hat{d}_{\CF}^{(n)}=\{s,\hat{x}_{\CF}^{(n)}\}$, we first derive the hidden representations, $z^{(n)}$ and $z_{\CF}^{(n)}$, respectively, by feeding them to the encoder $E_{\theta_2}$. Then, a discriminator $C_{\psi}$ is applied on $z^{(n)}$ and $z_{\CF}^{(n)}$ to predict whether the hidden representations are from observed or counterfactual samples, which is a binary classification task. We parameterize the discriminator $C_{\psi}$ by a multilayer neural network with the sigmoid function as the output layer and use the negative of the standard cross-entropy loss for binary classification tasks as the objective function to train the discriminator:
\begin{equation*}
    \mathcal{L}_{\textrm{C}}(\theta_2,\psi) = \frac{1}{N} \sum_{n=1}^N [\log (C_{\psi}(z^{(n)}))+\log(1-C_{\psi}(z_{\CF}^{(n)}))].
\end{equation*}
The discriminator is trained to accurately separate the hidden representations of observed and counterfactual samples. Meanwhile, to make the hidden representation derived from the encoder invariant to the change of sensitive attribute, the adversarial game is to train the encoder $E_{\theta_2}$ to fool the discriminator $C_{\psi}$ but still be good for reconstructing the original input. As a result, the objective function can be defined as a minimax problem:
\begin{equation}
\label{eq:objective}
    \min_{\theta_2,\phi_2}\max_{\psi} \mathcal{L}_{\textrm{AE}}(\theta_2,\phi_2) +\lambda\mathcal{L}_{\textrm{C}}(\theta_2,\psi),
\end{equation}
where $\lambda$ is a hyper-parameter to balance the reconstruction error and adversarial loss. Besides minimizing the reconstruction error $\mathcal{L}_{\textrm{AE}}$, the encoder also tries to maximize the cross-entropy loss for the discriminator $\mathcal{L}_{\textrm{C}}(\theta_2,\psi)$. Once the discriminator is unable to distinguish the hidden representations from factual or counterfactual data, we expect that both factual and counterfactual samples have similar reconstruction errors.

After training, the anomaly score for a new sample $d=\{s,x\}$ is computed based on the reconstruction error:
\begin{equation*}
    g(x)=\|x-D_{\phi_2} \circ E_{\theta_2}(x)\|_2^2.
\end{equation*}
If the anomaly score $g(x)>\tau$, where $\tau$ is a hyperparameter of the model, we label the sample as anomalous, i.e., $\hat{y}=1$.

\section{Experiments}

\subsection{Experimental Setup}

{\bf \noindent Datasets. } We conduct experiments on a synthetic dataset and two real-world datasets, Adult and COMPAS. Table \ref{tb:datasets} summarizes the statistics of three datasets.
\begin{table}[h]
\footnotesize \centering
\caption{Statistics of datasets.}
\label{tb:datasets}
\begin{tabular}{c|cc|cc|cc}
\hline
               & \multicolumn{2}{c|}{Synthetic}       & \multicolumn{2}{c|}{Adult}       & \multicolumn{2}{c}{COMPAS}     \\ \hline
               & \multicolumn{1}{c|}{Training} & Test & \multicolumn{1}{c|}{Training} & Test    & \multicolumn{1}{c}{Training} & Test\\ \hline
Normal (Y=0)   & \multicolumn{1}{c|}{12000}         & 4000     & \multicolumn{1}{c|}{12000}         & 4000     & \multicolumn{1}{c|}{2000}         & 1283\\ \hline
Abnormal (Y=1) & \multicolumn{1}{c|}{N/A}         & 400     & \multicolumn{1}{c|}{N/A}         & 800     & \multicolumn{1}{c|}{N/A}         & 384\\ \hline
\end{tabular}
\end{table}

\begin{figure}[!tbp]
  \centering
  \begin{minipage}[b]{0.23\textwidth}
    \centering
    \begin{subfigure}[t]{\textwidth}
    \includegraphics[width=0.98\textwidth]{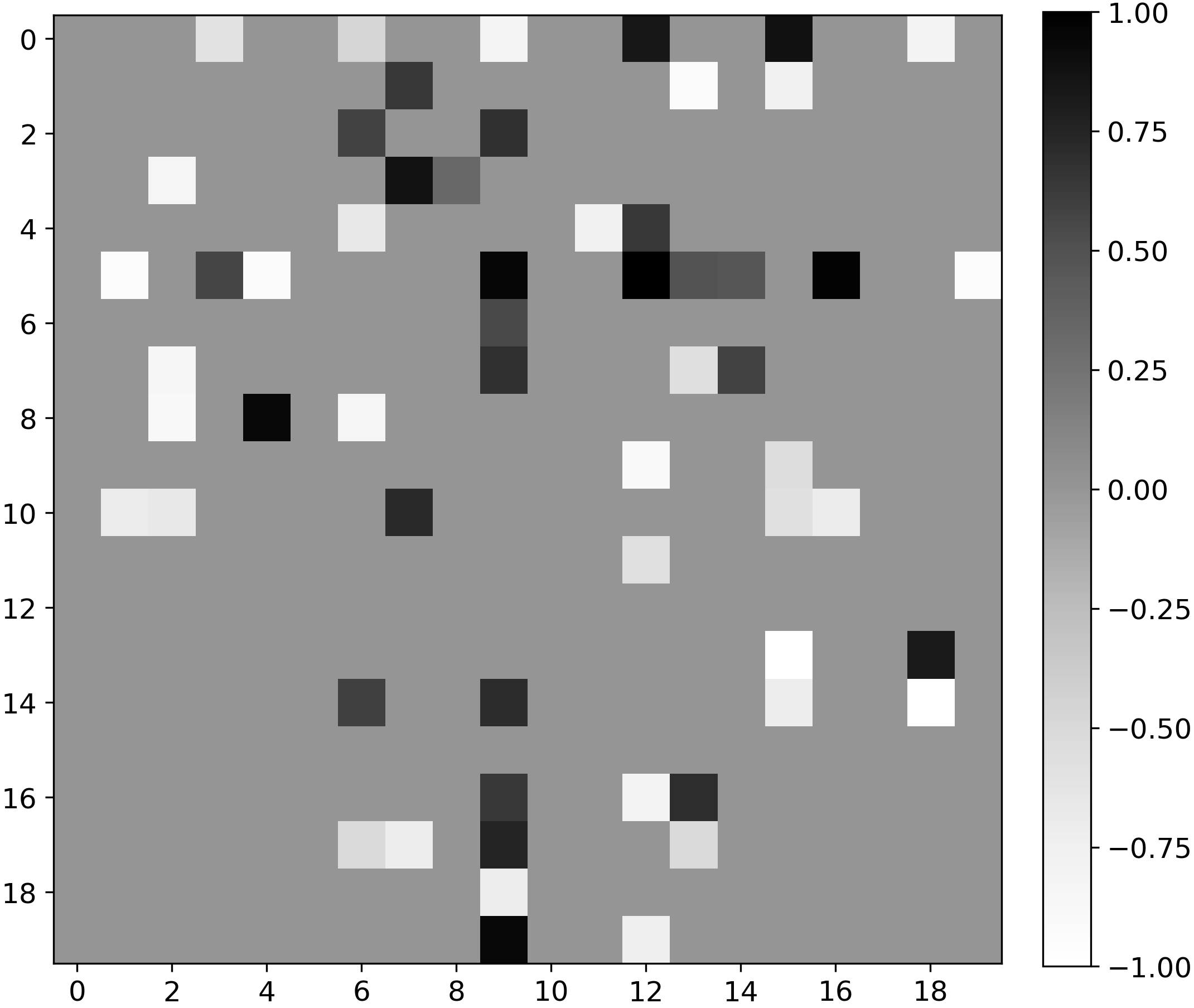}
    \end{subfigure}
    \caption{Adjacency matrix $A$}
    \label{fig:syn_A}
  \end{minipage}
  \hfill
  \begin{minipage}[b]{0.23\textwidth}
    \centering
    \begin{subfigure}[t]{\textwidth}
    \includegraphics[width=0.98\textwidth]{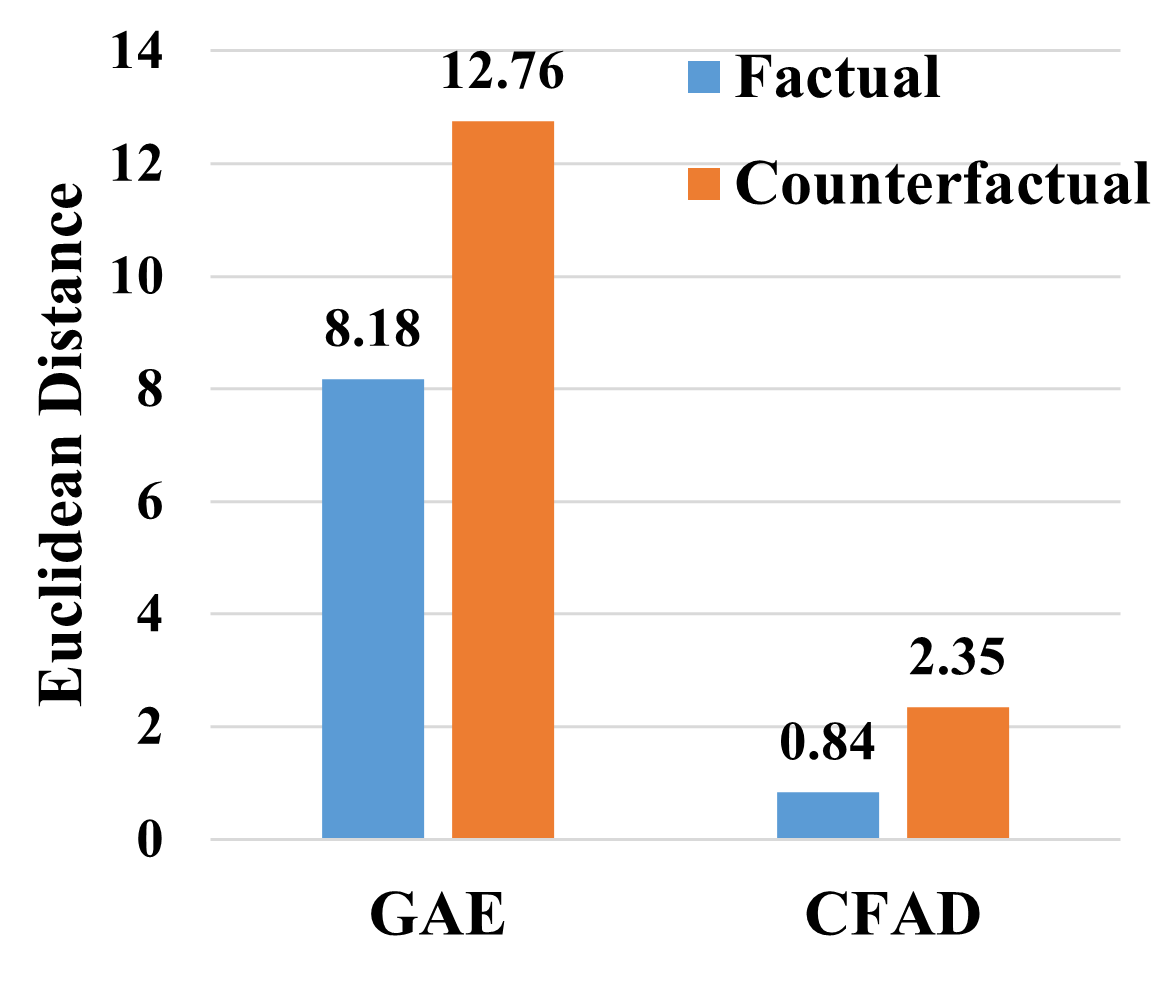}
    \end{subfigure}
    \caption{Results on data generation.}
    \label{fig:recon_error}
  \end{minipage}
  \hfill
  \begin{minipage}[b]{0.52\textwidth}
    \centering
    \begin{subfigure}[t]{.58\textwidth}
        \centering
      \includegraphics[width=0.98\textwidth]{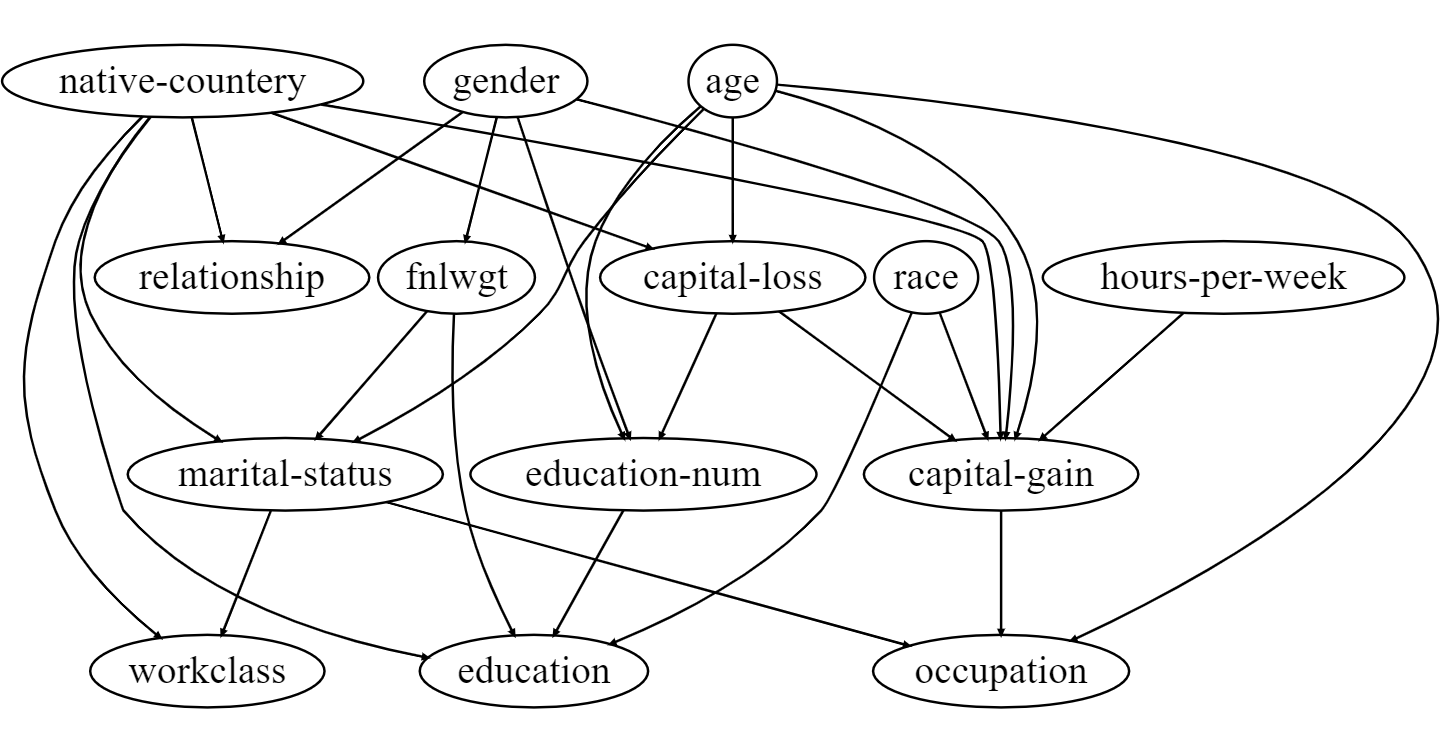}
      \caption{Adult}
      \label{fig:graph_adult}
    \end{subfigure}
    \hfill
    \begin{subfigure}[t]{0.32\textwidth}
      \centering
      \includegraphics[width=0.98\textwidth]{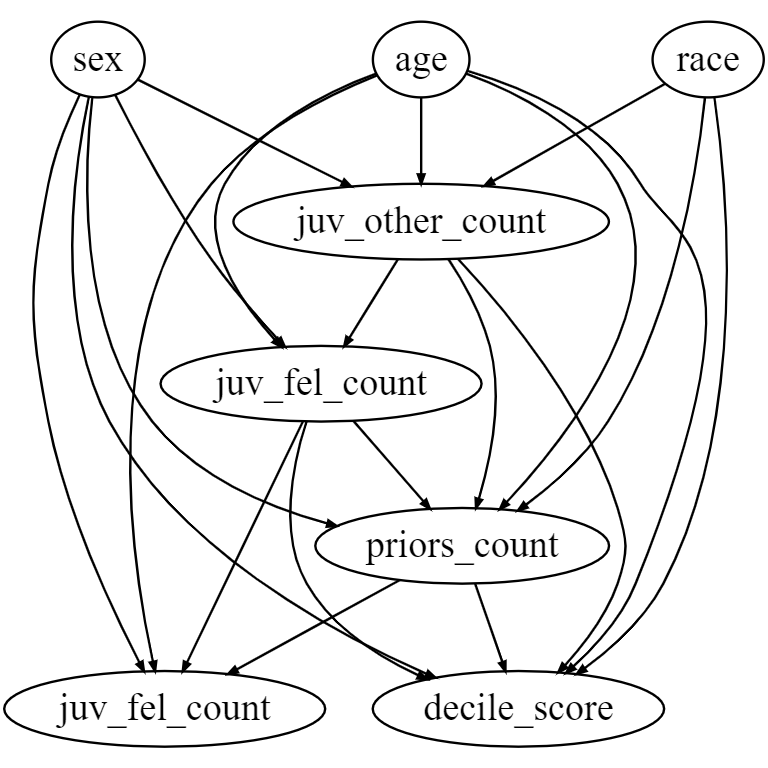}
      \caption{COMPAS}
      \label{fig:graph_compas}
    \end{subfigure}
    \caption{Learned causal graphs.}
    \label{fig:learned_graph}
  \end{minipage}
\end{figure}

\textbf{Synthetic Dataset.} We first build a synthetic dataset with 21 variables where we can obtain the ground truth of counterfactuals.
We first randomly generate the adjacency matrix $A$ of a causal graph using the Erd\H{o}s-R\'enyi model \cite{zheng2018dags} where one node is defined as a root node for representing the sensitive variable $S$. Figure \ref{fig:syn_A} shows the generated adjacency matrix $A$. The value of $S$ is randomly generated with binarized value \{-1, 1\} to indicate sensitive and non-sensitive groups. Then, similar to \cite{ngGraphAutoencoderApproach2019}, the rest 20 variables are generated based on the following data generating procedure: $X= 3 A^T \cos(X+1)+U$,
where $U$ is a standard Gaussian noise.
Finally, one leaf node is selected as the decision attribute $Y$ for determining anomalies. Specifically, for each sample, if the value of $Y$ is greater than $0.85$ quantile or smaller than $0.01$ quantile, we label this sample as an anomaly, i.e., $Y=1$. If the value of $Y$ is between $0.3$ and $0.7$ quantiles, we label the sample as normal, i.e., $Y=0$.
Meanwhile, for both training and test sets, for 50\% of the samples, their corresponding counterfactuals have labels that are different from the factual ones.

\textbf{Adult Dataset.} Adult is a real-world dataset with 14 features \cite{Dua:2019}. We treat ``\textit{gender}'' as the sensitive attribute and samples with ``\textit{income $>$ 50k}'' as anomalies. We normalize all continuous features and binarize all categorical features. Figure \ref{fig:graph_adult} shows the causal graph on Adult learned in Phase One of our approach. Note that we do not use the label ``income'' for training. Meanwhile, as we do not know the ground truth of counterfactuals, we use the generated counterfactual samples for measuring counterfactual fairness.

\textbf{COMPAS Dataset.} COMPAS is another real-world dataset \cite{dressel2018accuracy}, which consists of 8 features. We consider ``race'' as the sensitive attribute, where ``African-American'' and ``Caucasian'' are the disadvantage and advantage groups, respectively, and treat ``recidivists'' as anomalies. Similar to Adult, we normalize all continuous features and binarize all categorical features. Figure \ref{fig:graph_compas} shows the learned causal graph. 

{\bf \noindent Baselines.} 
We compare CFAD with the following baselines: 1) Principal Component Analysis (\textbf{PCA}), which is a dimensional reduction based anomaly detection approach; 2) One-class SVM (\textbf{OCSVM}), which is a one-class classification model that can detect outliers based on the observed normal samples; 3) Isolation Forest (\textbf{iForest}), which is a widely used tree-based anomaly detection model; 4) Autoencoder (\textbf{AE}), which is trained on normal data and widely-used for anomaly detection based on the deep autoencoder structure; 5) Deep Clustering based Fair Outlier Detection (\textbf{DCFOD}) \cite{songDeepClusteringBased2021}, which adopts the adversarial training to achieve the group fairness in anomaly detection; 6) Fairness-aware Outlier Detection (\textbf{FairOD}) \cite{shekharFairODFairnessawareOutlier2021}, which is also an autoencoder-based anomaly detection approach with fairness regularizers.

{\bf \noindent Evaluation Metrics.} We evaluate the performance of anomaly detection based on Area Under Precision-Recall Curve \textbf{(AUC-PR)}, Area Under Receiver Operating Characteristic Curve \textbf{(AUC-ROC)}, and \textbf{Macro-F1}. We evaluate counterfactual fairness by computing the \textbf{changing ratio} of the samples whose detection outcomes are different from those for their corresponding counterfactuals, i.e.,  
$changing\_ratio = \frac{\sum_{n=1}^{N} \mathbbm{1}[\hat{y}^{(n)}\neq \hat{y}^{(n)}_{\CF}]}{N},$
where $\mathbbm{1}[\cdot]$ is the indicator function.

{\bf \noindent Implementation Details.} 
Regarding baselines, we use Loglizer \cite{heexperience2016} to evaluate PCA, OC-SVM, and iForest. We implement FairOD and DCFOD based on public source code \cite{songDeepClusteringBased2021}. 
By default, the threshold $\tau$ for anomaly detection is set based on the $0.95$ quantile of reconstruction errors (AE, FairOD, and CFAD) or distance to the normal center (DCFOD) in the training set. Our code on CFAD is available online\footnote{\url{https://github.com/hanxiao0607/CFAD}}.

\subsection{Experimental Results}

{\bf \noindent Counterfactual Data Generation.}
Counterfactual data generation is a fundamental component of counterfactually fair anomaly detection. Therefore, we first evaluate the performance of counterfactual data generation in the synthetic dataset by comparing CFAD with  GAE \cite{ngGraphAutoencoderApproach2019} in terms of Euclidean distance between the generated and ground-truth samples. Figure \ref{fig:recon_error} shows the results. We can notice that on the factual data, CFAD achieves a much lower reconstruction error compared with GAE. More importantly, for counterfactual data generation, CFAD is much better compared with GAE. It indicates that by incorporating a variable-wise decoder $D_{\phi'_1}$ for data generation, CFAD can generate counterfactual samples with high fidelity.

\begin{table*}[ht!]
\small
\centering
\caption{Anomaly detection on synthetic and real datasets with threshold $\tau=0.95$. For AUC-PR, AUC-ROC and Macro-F1, the higher the value the better the effectiveness; for Changing Ratio, the lower the value the better the fairness.}
\label{tab:cf-res}
\resizebox{\textwidth}{!}{%
\begin{tabular}{c|cccc|cccc|cccc}
\hline
\multirow{2}{*}{Method} & \multicolumn{4}{c|}{Synthetic Dataset}                                                                                                       & \multicolumn{4}{c|}{Adult Dataset}                                                                                                           & \multicolumn{4}{c}{COMPAS Dataset}                                                                                                          \\ \cline{2-13} 
                        & \multicolumn{1}{c|}{AUC-PR}            & \multicolumn{1}{c|}{AUC-ROC}           & \multicolumn{1}{c|}{Macro-F1}          & Changing Ratio    & \multicolumn{1}{c|}{AUC-PR}            & \multicolumn{1}{c|}{AUC-ROC}           & \multicolumn{1}{c|}{Macro-F1}          & Changing Ratio    & \multicolumn{1}{c|}{AUC-PR}            & \multicolumn{1}{c|}{AUC-ROC}           & \multicolumn{1}{c|}{Macro-F1}          & Changing Ratio    \\ \hline
PCA                     & \multicolumn{1}{c|}{$0.992$} & \multicolumn{1}{c|}{$0.999$} & \multicolumn{1}{c|}{$0.908$} & $0.478$ & \multicolumn{1}{c|}{$0.238$} & \multicolumn{1}{c|}{$0.582$} & \multicolumn{1}{c|}{$0.476$} & $0.261$ & \multicolumn{1}{c|}{$0.365$} & \multicolumn{1}{c|}{$0.642$} & \multicolumn{1}{c|}{$0.595$} & $0.268$ \\ \hline
OC-SVM                  & \multicolumn{1}{c|}{$0.776$} & \multicolumn{1}{c|}{$0.953$} & \multicolumn{1}{c|}{$0.477$} & $0.399$ & \multicolumn{1}{c|}{$0.282$} & \multicolumn{1}{c|}{$0.638$} & \multicolumn{1}{c|}{$0.482$} & $0.285$ & \multicolumn{1}{c|}{$0.337$} & \multicolumn{1}{c|}{$0.593$} & \multicolumn{1}{c|}{$0.488$} & $0.376$ \\ \hline
iForest                 & \multicolumn{1}{c|}{$0.190$} & \multicolumn{1}{c|}{$0.693$} & \multicolumn{1}{c|}{$0.570$} & $0.271$ & \multicolumn{1}{c|}{$0.312$} & \multicolumn{1}{c|}{$0.658$} & \multicolumn{1}{c|}{$0.570$} & $0.279$ & \multicolumn{1}{c|}{$0.311$} & \multicolumn{1}{c|}{$0.567$} & \multicolumn{1}{c|}{$0.564$} & $0.415$ \\ \hline
AE                      & \multicolumn{1}{c|}{$0.957$} & \multicolumn{1}{c|}{$0.996$} & \multicolumn{1}{c|}{$0.883$} & $0.461$ & \multicolumn{1}{c|}{$0.349$} & \multicolumn{1}{c|}{$0.640$} & \multicolumn{1}{c|}{$0.608$} & $0.590$ & \multicolumn{1}{c|}{$0.344$} & \multicolumn{1}{c|}{$0.616$} & \multicolumn{1}{c|}{$0.581$} & $0.407$ \\ \hline
DCFOD                   & \multicolumn{1}{c|}{$0.383$} & \multicolumn{1}{c|}{$0.832$} & \multicolumn{1}{c|}{$0.721$} & $0.212$ & \multicolumn{1}{c|}{$0.249$} & \multicolumn{1}{c|}{$0.623$} & \multicolumn{1}{c|}{$0.533$} & $0.071$ & \multicolumn{1}{c|}{$0.260$} & \multicolumn{1}{c|}{$0.569$} & \multicolumn{1}{c|}{$0.466$} & $0.067$ \\ \hline
FairOD                  & \multicolumn{1}{c|}{$0.580$} & \multicolumn{1}{c|}{$0.873$} & \multicolumn{1}{c|}{$0.689$} & $0.261$ & \multicolumn{1}{c|}{$0.222$} & \multicolumn{1}{c|}{$0.621$} & \multicolumn{1}{c|}{$0.531$} & $0.131$ & \multicolumn{1}{c|}{$0.265$} & \multicolumn{1}{c|}{$0.548$} & \multicolumn{1}{c|}{$0.493$} & $0.068$ \\ \hhline{|=|====|====|====|} 
CFAD                    & \multicolumn{1}{c|}{$0.947$} & \multicolumn{1}{c|}{$0.996$} & \multicolumn{1}{c|}{$0.930$} & $0.199$ & \multicolumn{1}{c|}{$0.319$} & \multicolumn{1}{c|}{$0.589$} & \multicolumn{1}{c|}{$0.576$} & $0.057$ & \multicolumn{1}{c|}{$0.314$} & \multicolumn{1}{c|}{$0.596$} & \multicolumn{1}{c|}{$0.539$} & $0.049$ \\ \hline

\end{tabular}%
}
\end{table*}

{\bf \noindent Anomaly Detection.}
We further evaluate the performance of anomaly detection in terms of effectiveness as well as fairness. Table \ref{tab:cf-res} shows the evaluation results. We report the mean value after five runs.

\textit{Synthetic Dataset.} CFAD can well balance the effectiveness and fairness in anomaly detection with high AUC-PR, AUC-ROC, and Macro-F1 values and a low changing ratio. As a classical deep anomaly detection model, AE can achieve high AUC-PR, AUC-ROC, and Macro-F1 values. However, its changing ratio is high meaning that AE cannot ensure counterfactual fairness. For other baselines, PCA can achieve the best performance for anomaly detection in terms of AUC-PR and AUC-ROC, but has the highest changing ratio. For iForest and OC-SVM, they can neither achieve good anomaly detection performance nor good fairness. Finally, DCFOD and FairOD, which achieve group fairness in anomaly detection, both have relatively low changing ratios, but their effectiveness in anomaly detection is not satisfactory.

\textit{Real Datasets.} We have similar observations on the Adult and COMPAS datasets. CFAD achieves good performance on both effectiveness and fairness. For baselines that have no fairness component, their performance is good in terms of the effectiveness in anomaly detection, but they all have high changing ratios. Similarly, although DCFOD and FairOD have relatively low changing ratios, their effectiveness is much worse than other approaches.

\begin{figure*}[h!]
\centering
    \begin{subfigure}{.3\textwidth}
      \centering
      \includegraphics[width=0.98\textwidth]{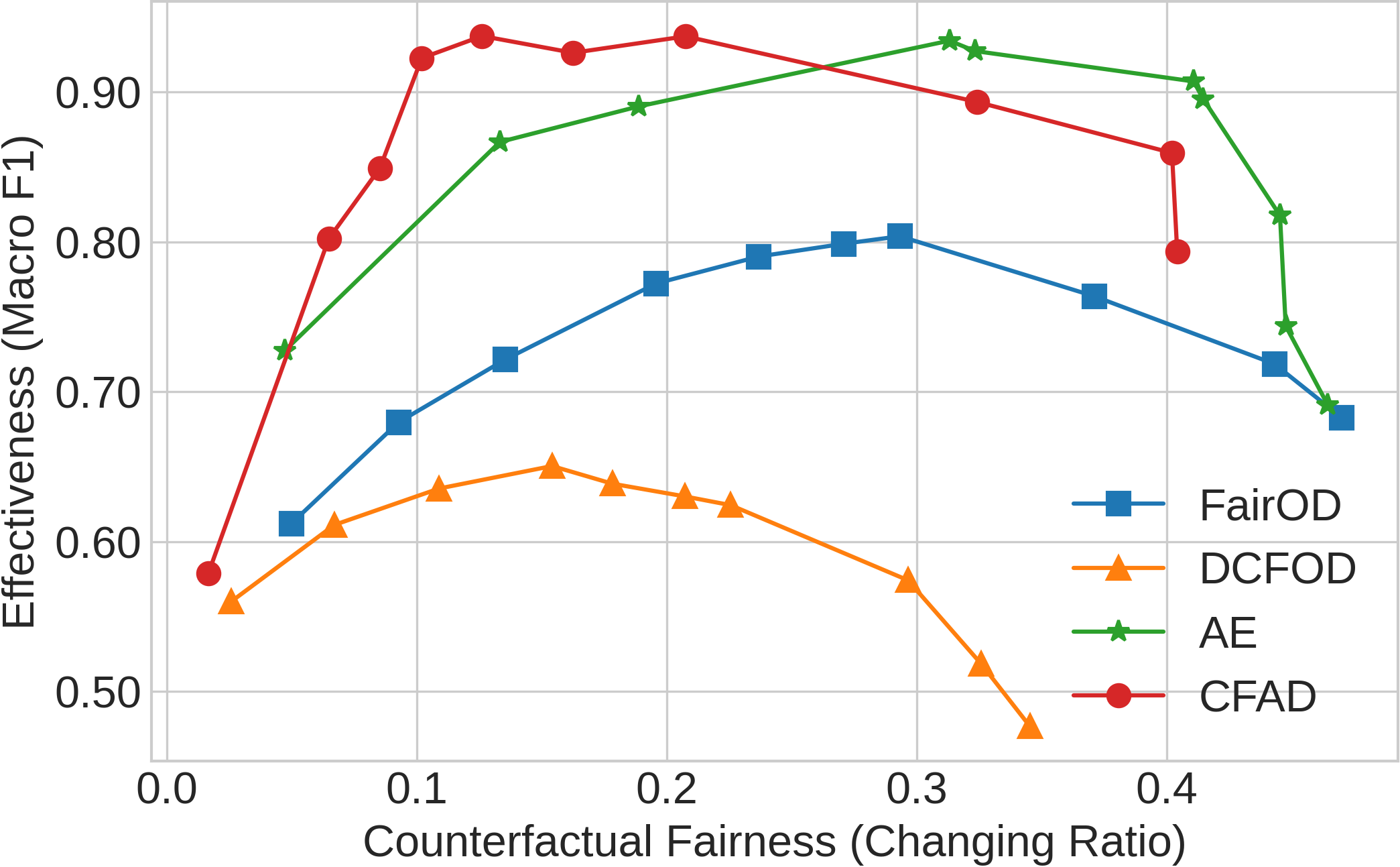}
      \caption{Synthetic}
      \label{fig:syn_vs}
    \end{subfigure}%
    \hfill
    \begin{subfigure}{.3\textwidth}
      \centering
      \includegraphics[width=0.98\textwidth]{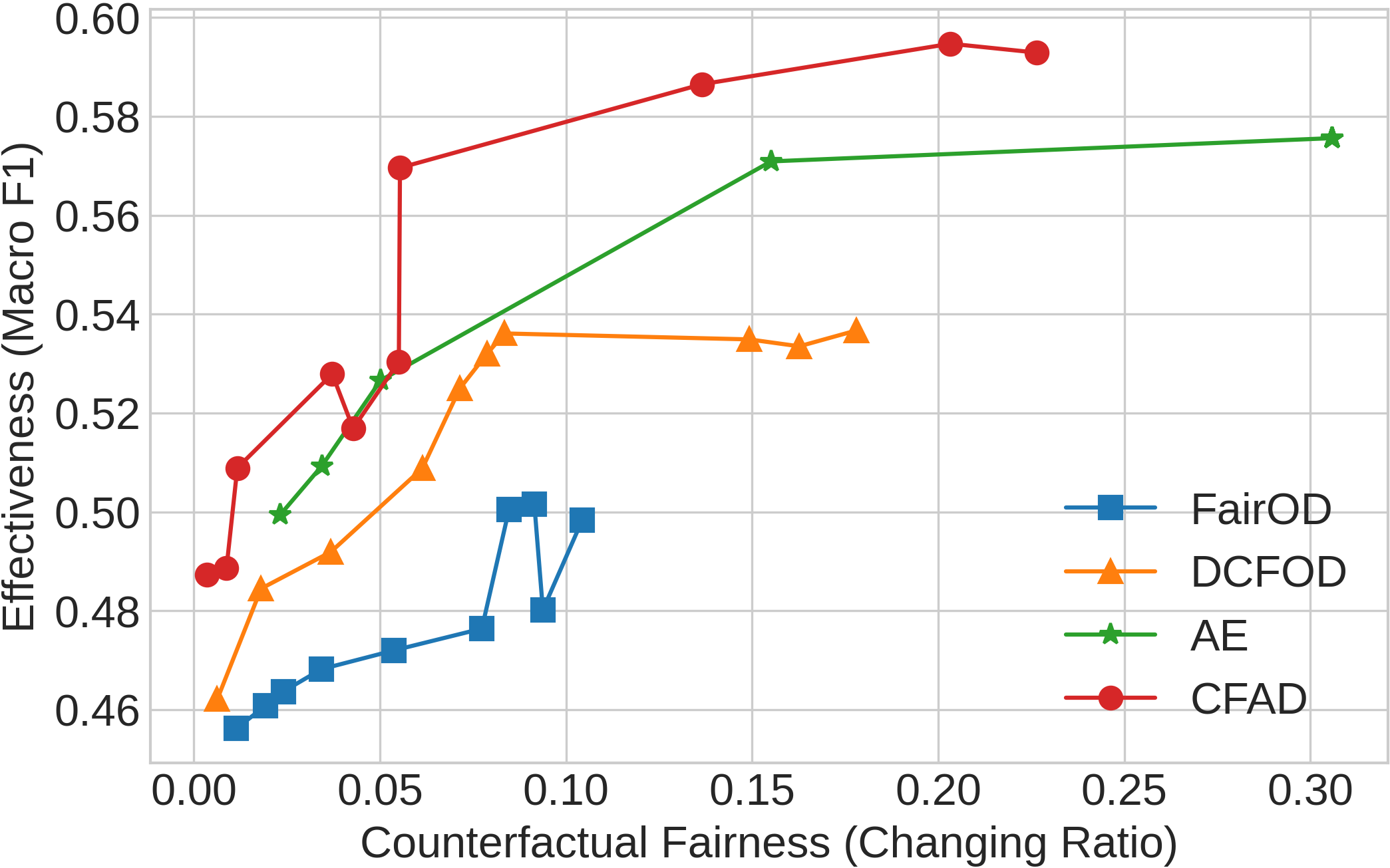}
      \caption{Adult}
      \label{fig:adult_vs}
    \end{subfigure}
    \hfill
    \begin{subfigure}{.3\textwidth}
      \centering
      \includegraphics[width=0.98\textwidth]{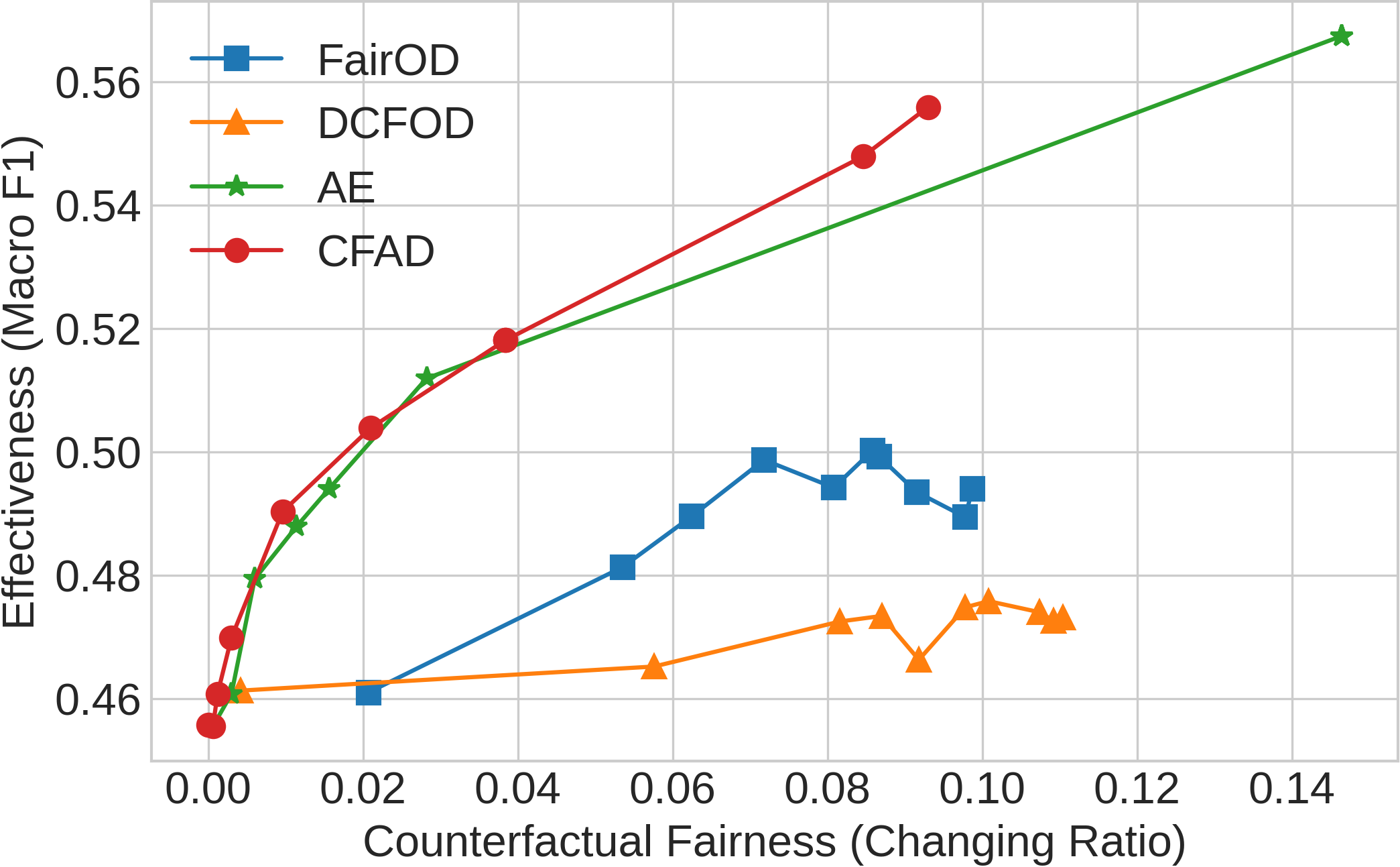}
      \caption{COMPAS}
      \label{fig:compas_vs}
    \end{subfigure}
\caption{Trade-off between effectiveness and fairness.}
\label{fig:vs}
\end{figure*}

{\bf \noindent Trade-off Between Effectiveness and Fairness.}
We further investigate the trade-off between effectiveness and fairness by varying the threshold as different quantiles of reconstruction errors or distances in the training set. We plot the effectiveness and fairness of each threshold setting of four approaches CFAD, AE, DCFOD, and FairOD in Figure \ref{fig:vs}, where the x-axis is the changing ratio (counterfactual fairness), the y-axis indicates the Macro-F1 score (effectiveness), and each dot in the line indicates the result from one threshold. The dots from right to left indicate the performance based on quantiles including $\{0.8,0.85,0.9,0.95,0.97,0.98,0.99,0.995,0.999\}$. Ideally, we expect an anomaly detection model can achieve a high Marco-F1 score with a low changing ratio, which is the top left corner of the figure. 

As shown in Figure \ref{fig:vs}, CFAD performs best when the effectiveness trades off with fairness, as CFAD is closest to the top left corner of the figure. Specifically, on the Synthetic dataset, CFAD achieves much higher Macro-F1 values (effectiveness) with similar changing rates (fairness) compared with DCFOD and FairOD.  Meanwhile, for most of the thresholds chosen based on quantiles, CFAD has higher Macro-F1 and lower changing ratios compared with AE. On the Adult and COMPAS datasets, CFAD can have higher Macro-F1 values and lower changing ratios compared with DCFOD and FairOD. AE can also have reasonable performance in terms of effectiveness on real datasets, but it cannot achieve fairness with large changing ratios.

\begin{figure*}[h!]
\centering
    \begin{subfigure}[c]{.24\textwidth}
      \centering
    \includegraphics[width=0.98\textwidth]{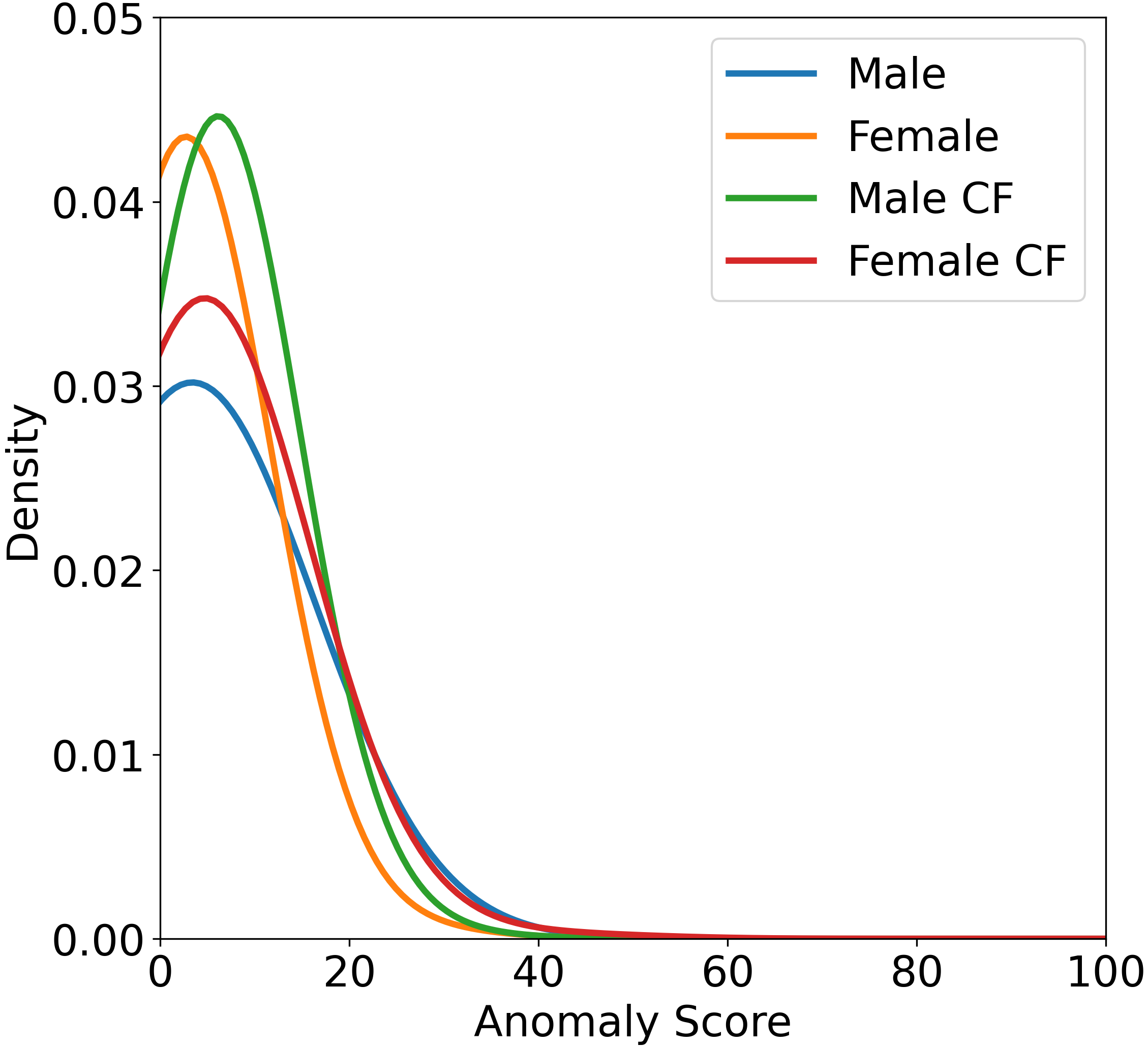}
      \caption{AE-Adult}
      \label{fig:adult_ae}
    \end{subfigure}%
    \hfill
    \begin{subfigure}[c]{.24\textwidth}
      \centering
      \includegraphics[width=0.98\textwidth]{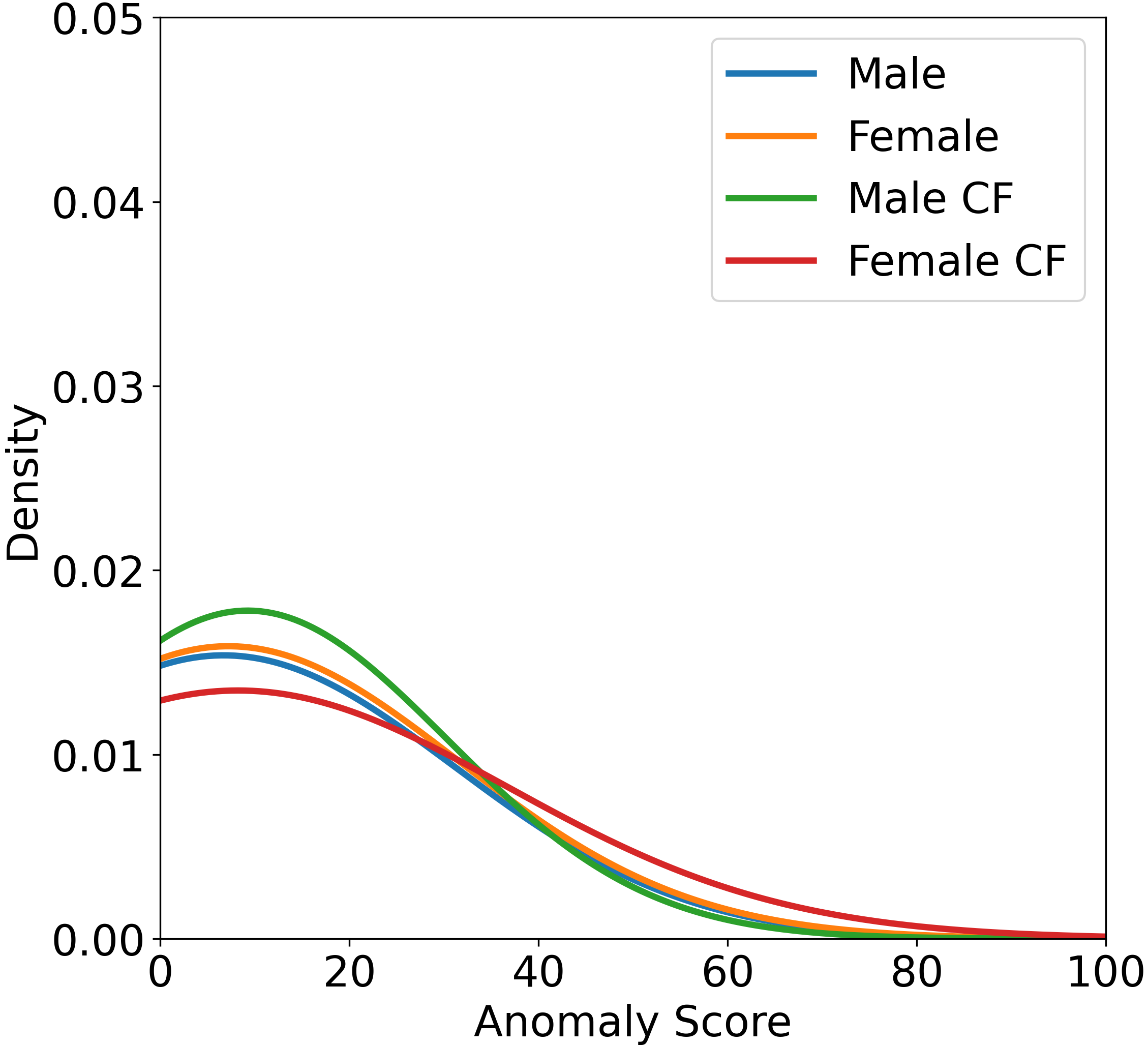}
      \caption{CFAD-Adult}
      \label{fig:adult_cfad}
    \end{subfigure}
    \hfill
    \begin{subfigure}[c]{.24\textwidth}
      \centering
      \includegraphics[width=0.98\textwidth]{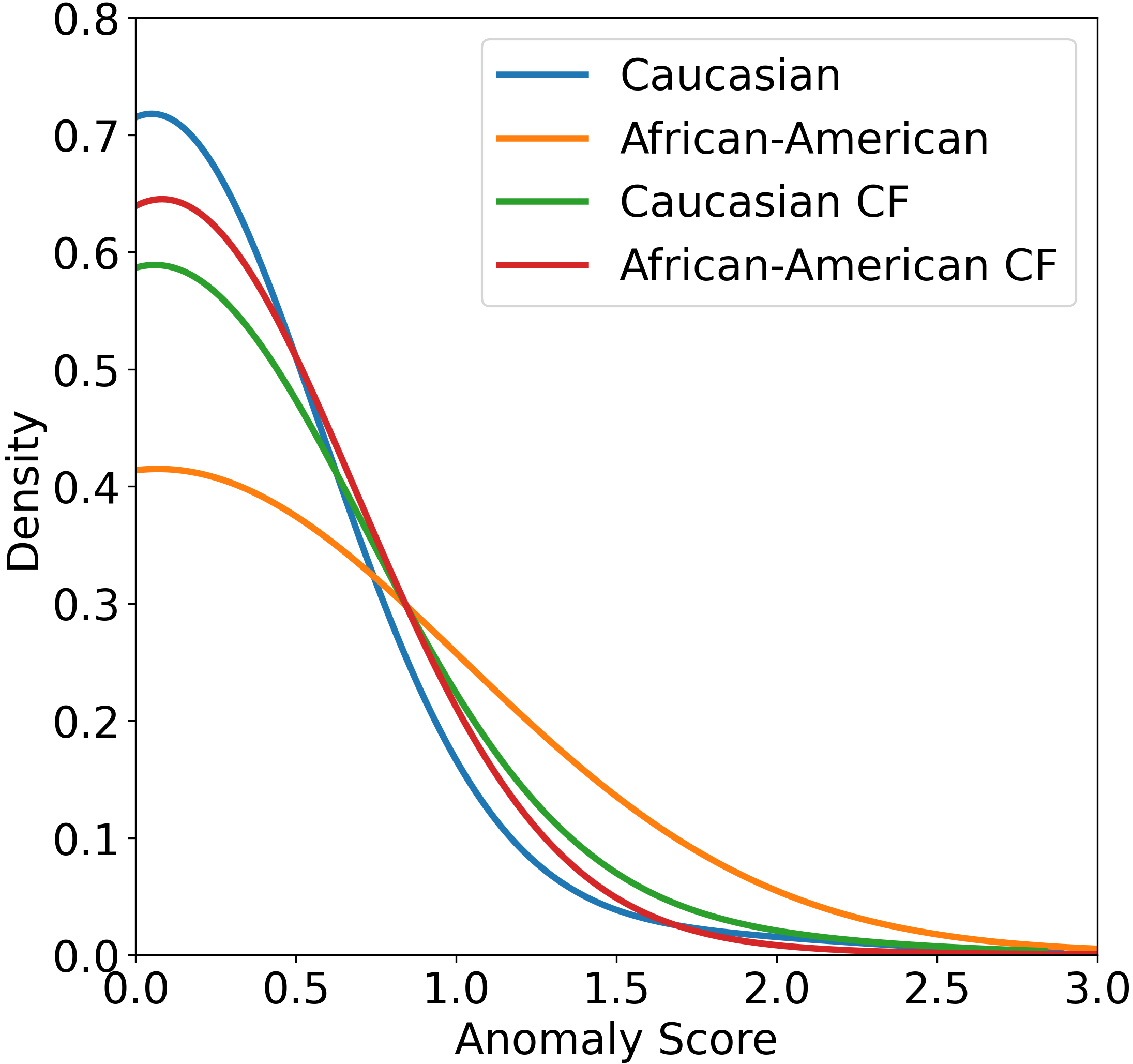}
      \caption{AE-COMPAS}
      \label{fig:compas_ae}
    \end{subfigure}%
    \hfill
    \begin{subfigure}[c]{.24\textwidth}
      \centering
      \includegraphics[width=0.98\textwidth]{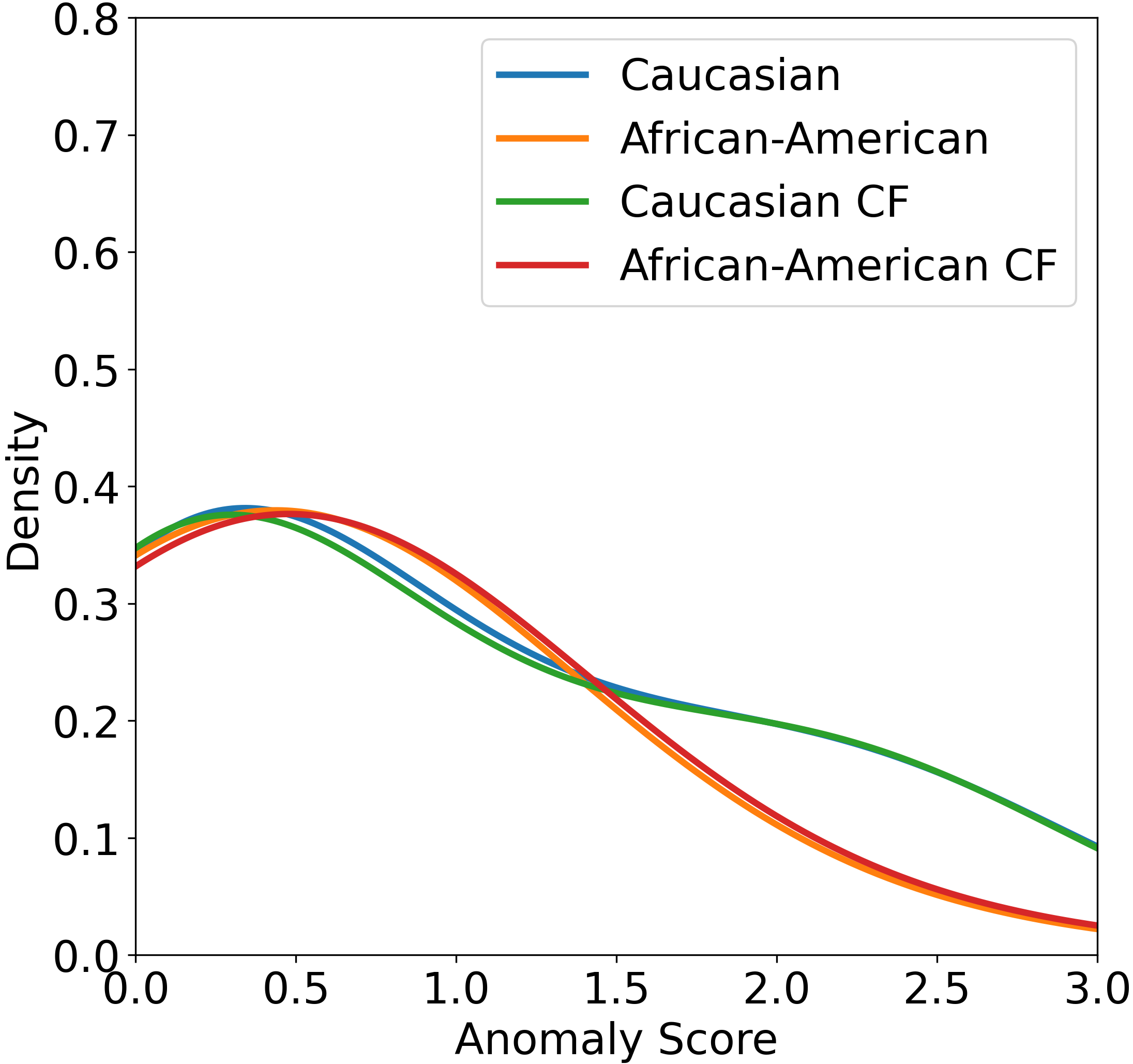}
      \caption{CFAD-COMPAS}
      \label{fig:compas_cfad}
    \end{subfigure}
\caption{AE vs. CFAD.}
\label{fig:ae_vs_cfad}
\end{figure*}

{\bf \noindent AE vs CFAD.}
CFAD improves the vanilla AE-based anomaly detection model with counterfactual fairness. We further conduct a fine-grained comparison between AE and CFAD by analyzing the density distributions of anomaly scores over advantage, disadvantage, counterfactual advantage, and counterfactual disadvantage groups on two real-world datasets, Adult and COMPAS, shown in Figure \ref{fig:ae_vs_cfad}. On the Adult dataset, we can observe that there is a big gap between factual and counterfactual samples (Male v.s. Male CF and Female v.s. Female CF) in terms of the anomaly score density distribution from AE (Figure \ref{fig:adult_ae}). In contrast, the gaps from CFAD are much smaller (Figure \ref{fig:adult_cfad}). It means the anomaly scores derived from CFAD are much more consistent on factual and counterfactual samples, which ensures counterfactual fairness. The advantage of CFAD on the COMPAS dataset is also very clear as the anomaly score density distributions from factual and counterfactual samples are very close (Figure \ref{fig:compas_cfad}).

\begin{figure*}[h]
\centering
    \begin{subfigure}{0.43\textwidth}
      \centering
      \includegraphics[width=0.98\textwidth]{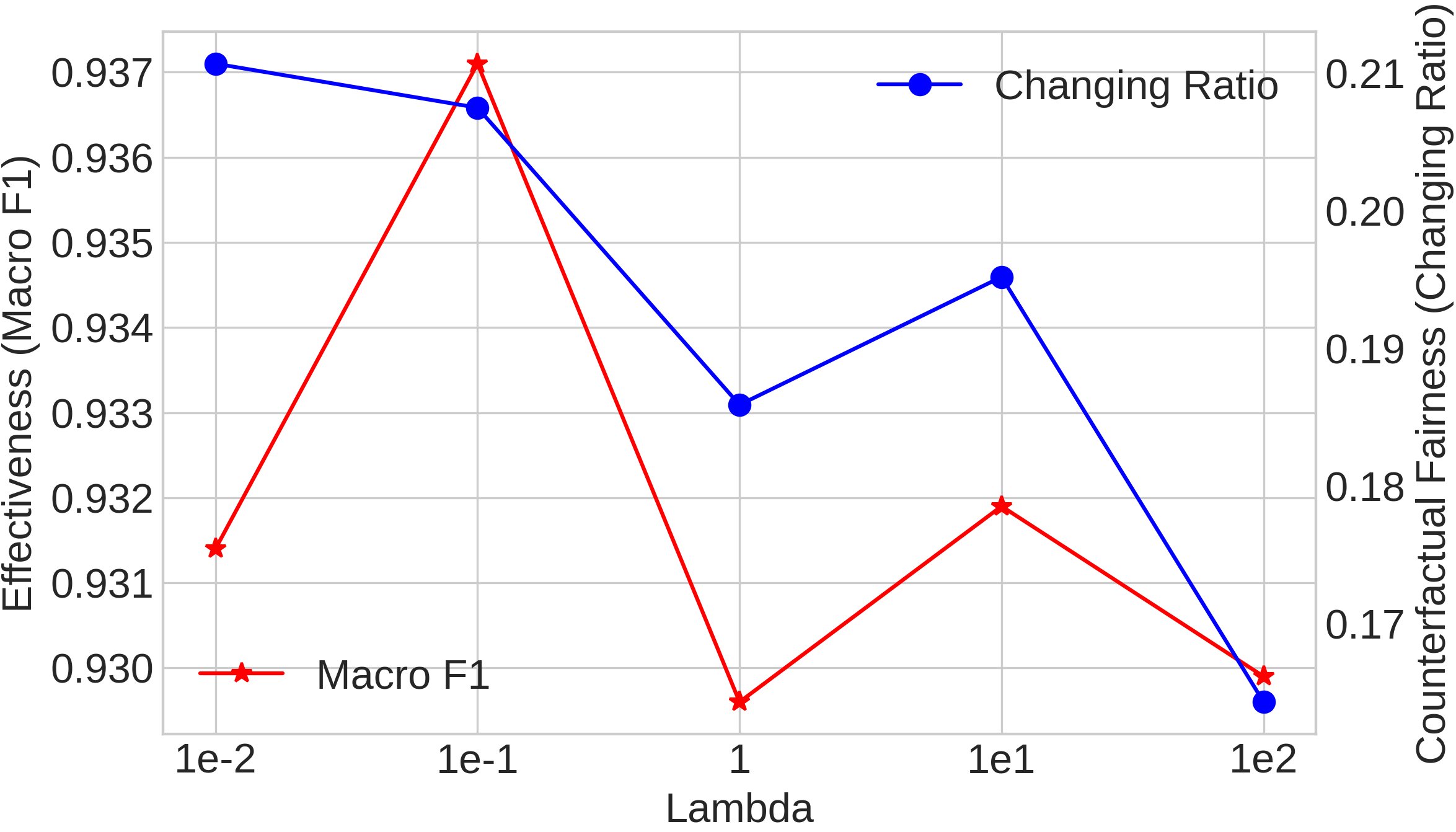}
      \caption{$\lambda$ in Eq.~\eqref{eq:objective}}
      \label{fig:syn_lambda}
    \end{subfigure}%
    \hfill
    \begin{subfigure}{0.43\textwidth}
      \centering
      \includegraphics[width=0.98\textwidth]{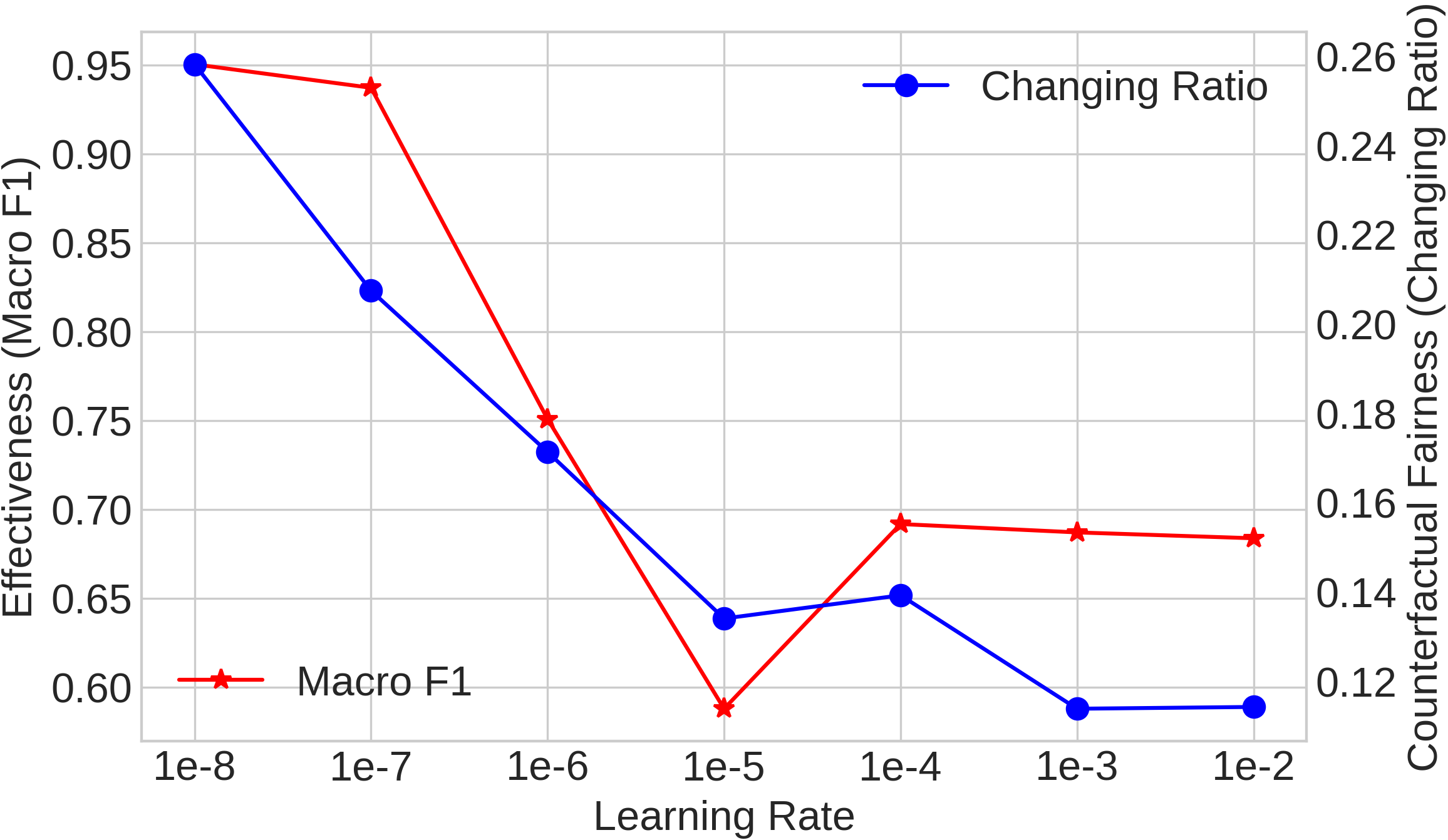}
      \caption{learning rate}
      \label{fig:syn_lr}
    \end{subfigure}
\caption{Sensitivity analysis.}
\label{fig:sa}
\end{figure*}
{\bf \noindent Sensitivity Analysis.}
In Phase Two of CFAD, the balance between effectiveness and fairness is controlled by two hyper-parameters, the $\lambda$ in Eq.~\eqref{eq:objective} and the learning rate to update the autoencoder $D_{\phi_2}\circ E_{\theta_2}$. We investigate how these two hyper-parameters affect the performance of CFAD. We conduct sensitivity analysis on the synthetic dataset because only synthetic data have the ground truth information about counterfactuals.
Figure \ref{fig:syn_lambda} shows that setting a large $\lambda$ by emphasizing the fairness term can reduce the changing ratio, i.e., improving the fairness, which meets the expectation. Meanwhile, we can notice that tuning $\lambda$ does not significantly change the Macro-F1, which means the effectiveness of the anomaly detection model is not heavily influenced by $\lambda$. Figure \ref{fig:syn_lr} shows the performance changes when we tune the learning rate for fine-tuning the autoencoder. We can notice that tuning the learning rate can significantly impact both Macro-F1 and changing ratio. This is because a large learning rate in the fine-tuning phase for updating the autoencoder can make the representations of observed and counterfactual data close to each other driven by the minmax game. As a result, the model will be fairer with a lower changing ratio. On the other hand, the performance of anomaly detection will get damaged due to the potential anomalies in the counterfactual data. Therefore, setting a small learning rate to fine-tune the autoencoder is important to well-balance effectiveness and fairness.

\section{Conclusions}
In this work, we have developed a counterfactually fair anomaly detection (CFAD) framework, which is able to effectively detect anomalies and also ensure counterfactual fairness. The core idea of CFAD is to generate counterfactual data governed by a learned causal structure based on the proposed graph autoencoder model. Then, by using a vanilla autoencoder as the anomaly detection model, an adversarial training strategy is adopted to ensure the representations derived by the autoencoder without the information of sensitive attributes. After that, counterfactual fairness is achieved by having similar reconstruction errors for both factual and counterfactual samples. 
The experimental results show that CFAD can achieve counterfactually fair anomaly detection while well-balancing the trade-off between effectiveness and fairness.

\section*{Acknowledgement}
This work was supported in part by NSF 1910284 and 2103829.

\end{document}